\documentclass[letterpaper]{article} 
\usepackage{aaai24} 
\usepackage{times}  
\usepackage{helvet}  
\usepackage{courier}  
\usepackage[hyphens]{url}  
\usepackage{graphicx} 
\usepackage{natbib}  
\usepackage{caption} 
\usepackage{algorithm}
\usepackage{algorithmic}
\usepackage{amsmath}
\usepackage{amsfonts} 
\usepackage{newfloat}
\usepackage{listings}
\usepackage{adjustbox}

\usepackage{xspace}
\usepackage{multirow}
\usepackage[table,xcdraw]{xcolor}
\usepackage[T1]{fontenc}

\DeclareCaptionStyle{ruled}{labelfont=normalfont,labelsep=colon,strut=off} 
\floatstyle{ruled}
\newfloat{listing}{tb}{lst}{}
\floatname{listing}{Listing}
\makeatletter
\DeclareRobustCommand\onedot{\futurelet\@let@token\@onedot}
\def\@onedot{\ifx\@let@token.\else.\null\fi\xspace}

\def\etc{\emph{etc}\onedot} 
 
\def\etal{\emph{et al}\onedot}
\makeatother

\title{Devignet: High-Resolution Vignetting Removal via a Dual Aggregated Fusion Transformer With Adaptive Channel Expansion}

\author {
    Shenghong Luo\textsuperscript{\rm 1}\equalcontrib,
    Xuhang Chen\textsuperscript{\rm 123}\equalcontrib,
    Weiwen Chen\textsuperscript{\rm 12},
    Zinuo Li\textsuperscript{\rm 12},\\
    Shuqiang Wang\textsuperscript{\rm 2}\thanks{Corresponding Author},
    Chi-Man Pun\textsuperscript{\rm 1}\footnotemark[2]
}
\affiliations {
    \textsuperscript{\rm 1}University of Macau\\
    \textsuperscript{\rm 2}Shenzhen Institutes of Advanced Technology, Chinese Academy of Sciences\\
    \textsuperscript{\rm 3}Huizhou University\\
}

\begin{document}

\maketitle

\begin{abstract}


Vignetting commonly occurs as a degradation in images resulting from factors such as lens design, improper lens hood usage, and limitations in camera sensors. This degradation affects image details, color accuracy, and presents challenges in computational photography. Existing vignetting removal algorithms predominantly rely on ideal physics assumptions and hand-crafted parameters, resulting in the ineffective removal of irregular vignetting and suboptimal results. Moreover, the substantial lack of real-world vignetting datasets hinders the objective and comprehensive evaluation of vignetting removal. To address these challenges, we present Vigset, a pioneering dataset for vignetting removal. Vigset includes 983 pairs of both vignetting and vignetting-free high-resolution ($5340\times3697$) real-world images under various conditions. In addition, We introduce DeVigNet, a novel frequency-aware Transformer architecture designed for vignetting removal. Through the Laplacian Pyramid decomposition, we propose the Dual Aggregated Fusion Transformer to handle global features and remove vignetting in the low-frequency domain. Additionally, we propose the Adaptive Channel Expansion Module to enhance details in the high-frequency domain. The experiments demonstrate that the proposed model outperforms existing state-of-the-art methods. The code, models, and dataset are available at \url{https://github.com/CXH-Research/DeVigNet}.

\end{abstract}

\section{Introduction}
\begin{figure}[t]
    \begin{minipage}[b]{1.0\linewidth}
        \begin{minipage}[b]{.32\linewidth}
            \centering
            \centerline{\includegraphics[width=\linewidth]{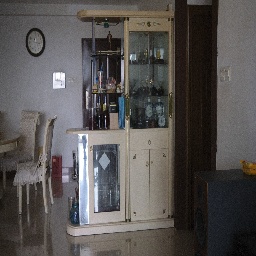}}
            \centerline{(a) Input}\medskip
        \end{minipage}
        \hfill
        \begin{minipage}[b]{.32\linewidth}
            \centering
            \centerline{\includegraphics[width=\linewidth]{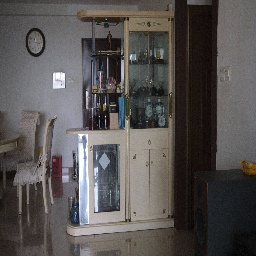}}
            \centerline{(b) RIVC}\medskip
        \end{minipage}
        \hfill
        \begin{minipage}[b]{0.32\linewidth}
            \centering
            \centerline{\includegraphics[width=\linewidth]{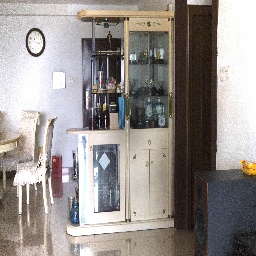}}
            \centerline{(c) SIVC}\medskip
        \end{minipage}
    \end{minipage}
    \begin{minipage}[b]{1.0\linewidth}
        \begin{minipage}[b]{.32\linewidth}
            \centering
            \centerline{\includegraphics[width=\linewidth]{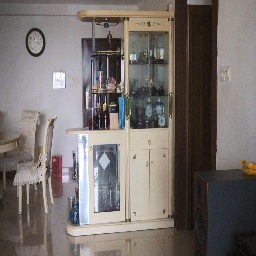}}
            \centerline{(d) ELGAN}\medskip
        \end{minipage}
        \hfill
        \begin{minipage}[b]{.32\linewidth}
            \centering
            \centerline{\includegraphics[width=\linewidth]{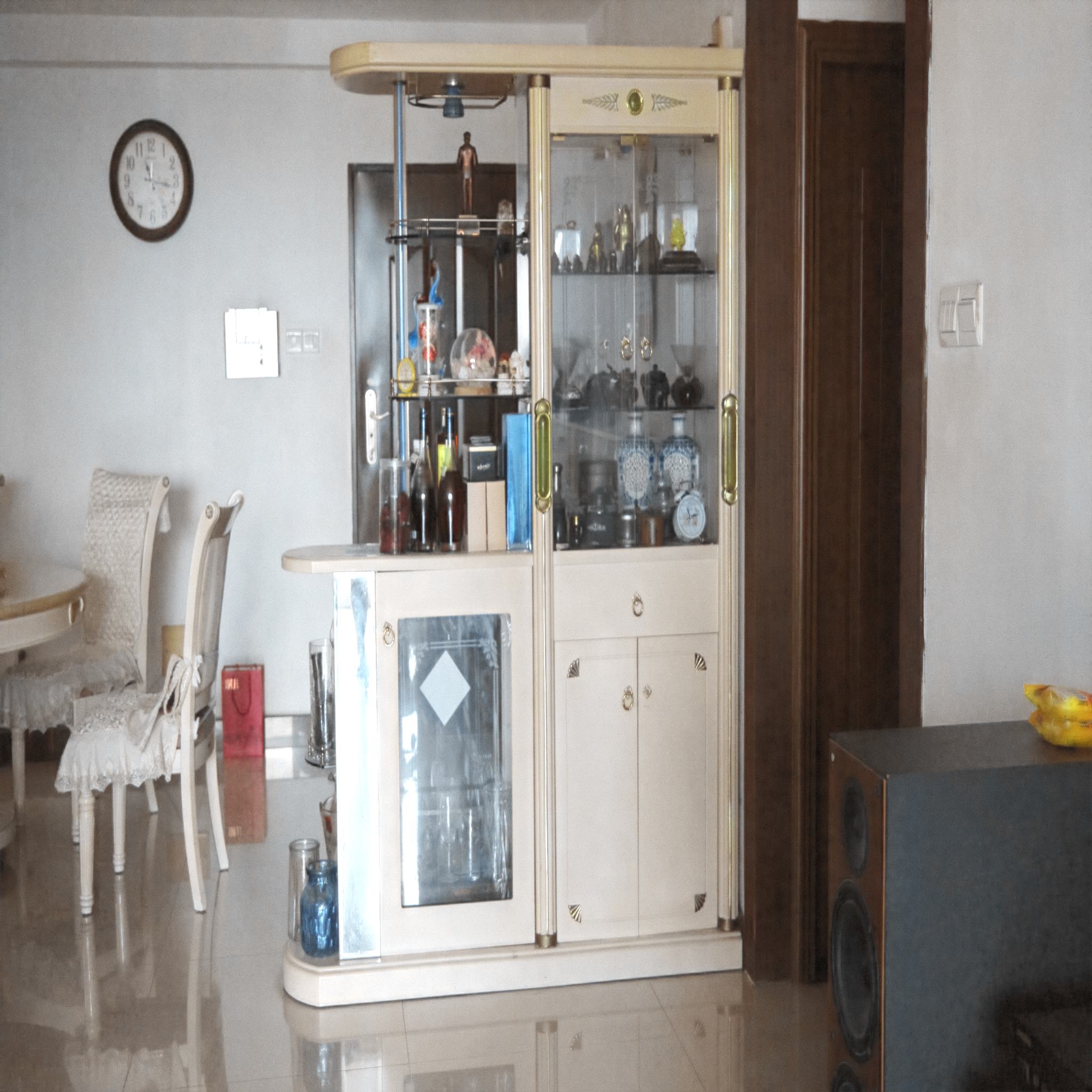}}
            \centerline{(e) Ours}\medskip
        \end{minipage}
        \hfill
        \begin{minipage}[b]{0.32\linewidth}
            \centering
            \centerline{\includegraphics[width=\linewidth]{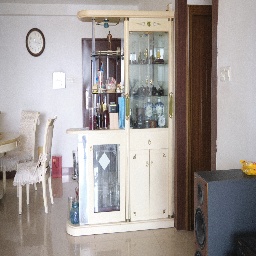}}
            \centerline{(f) Target}\medskip
        \end{minipage}
    \end{minipage}
    \caption{
    The visual results display the effects of different methods on the input (a) with vignetting, including the traditional methods (b) and (c), the LLIE model (d), and DevigNet (e). Our model effectively removes vignetting.
    } 
    \label{fig:teaser}
\end{figure}

Vignetting is a common optical degradation that results in a gradual decrease in brightness toward the edges of an image. It occurs due to multiple factors such as lens characteristics, filter presence, aperture settings, focal length settings, \etc.	

Some may confuse the difference between Low-Light Image Enhancement (LLIE) and vignetting removal. LLIE focuses on enhancing the overall brightness of images captured in low-light conditions. Its goal is to improve visibility, reduce noise, and enhance contrast in dark regions. On the other hand, vignetting removal specifically addresses the uneven light projection effects in specific regions of an image, typically towards the edges. Its purpose is to correct this effect and restore a more uniform brightness across the image. Therefore, these two tasks serve distinct purposes and aim to enhance different aspects of image quality.	




There are mathematical and prior-based methods available for vignetting removal~\cite{zheng2008single,2013Single,lopez2015revisiting}. Nevertheless, these approaches have limitations. These approaches ideally assume that the optical center is located at the center of the image, which may not be valid in real-world scenarios. Moreover, these methods can demonstrate bias under certain conditions and frequently necessitate extensive parameter adjustments to achieve optimal performance. In addition, these parameters are highly sensitive to high-resolution images, often leading to inferior outcomes. Another significant challenge arises from the absence of ground truth in the datasets used for evaluation, which contributes to subjective assessments of the experimental results. 

To tackle the issue of vignetting, we introduce a dataset named VigSet. The dataset comprises 983 pairs of images captured under various environmental and lighting conditions. Each pair consists of an image captured under optimal lighting conditions, without vignetting, and its corresponding vignetting-free ground truth image. Additionally, we present DeVigNet, a network that employs Dual Aggregated Fusion Transformer and Adaptive Channel Expansion for vignetting removal.
We utilize the Laplacian Pyramid~\cite{burt1987laplacian,liang2021high,Li_2023_ICCV,ijcai2023p129} to decompose the image into high-frequency and low-frequency components~\cite{you2022fine}. This decomposition enhances vignetting removal by image sharpening and edge enhancement. We introduce the Dual Aggregated Fusion Transformer for vignetting removal in the low-frequency component, as it primarily contains smoother color information that represents the overall color and distribution. Additionally, the high-frequency component utilizes the Adaptive Channel Expansion Module for handling color edges, texture details, and specific color characteristics. Our experimental results demonstrate the state-of-the-art performance of our method. In addition, Figure~\ref{fig:teaser} provides an intuitive indication of the favorable outcomes attained by DeVigNet.
The main contributions of this article are as follows:
\begin{itemize}
\item We present VigSet, the first vignetting dataset that includes high-resolution vignetting images along with the corresponding vignetting-free ground truth. VigSet aims to alleviate the current scarcity of vignetting datasets by providing a substantial number of samples accompanied by accurate ground truth information.
\item We propose DeVigNet, a network that is based on the Dual Aggregated Fusion Transformer and Adaptive Channel Expansion. It represents the first learning-based model for high-resolution vignetting removal.
\item Quantitative and qualitative experiments demonstrate that DeVigNet outperforms state-of-the-art methods on vignetting datasets.
\end{itemize}

\section{Related Work}

\subsection{Vignetting Removal}

A limited number of studies in the field of traditional vignetting removal has proposed methods that are based on mathematical principles, statistical analysis, and prior knowledge. SIVC~\cite{2013Single} utilizes the symmetry properties of semicircular tangential gradients and RG distributions to estimate the optical center and correct vignetting. Goldman \etal assume that the vignetting in the image exhibits radial symmetry around its center~\cite{goldman2010vignette}. RIVC~\cite{lopez2015revisiting}, addresses vignetting removal through the minimization of the log-intensity entropy.


\subsection{Low-Light Image Enhancement}
Traditional methods~\cite{6512558,guo2016lime,597272,cai2017joint,fu2016weighted,fu2015probabilistic} for Low-Light Image Enhancement often refers to the Retinex theory or histogram equalization. 

Recently, the utilization of these learning-based approaches gain traction as a prevalent solution for enhancing low-light images. Several widely recognized datasets are utilized by these learning-based methods. For instance, 
Wei \etal propose a Low-Light dataset (LOL)~\cite{wei2018deep} containing pairs of low/normal-light images.
The MIT-Adobe FiveK dataset~\cite{bychkovsky2011learning} comprises 5,000 photos that are manually annotated. To ensure the highest quality, the dataset underwent retouching by a team of 5 trained photographers, rendering it well-suited for supervised learning in the context of LLIE.
In terms of effective methods, certain technologies have also made contributions.
Wang \etal propose Uformer~\cite{wang2021uformer} employing both local and global dependencies to restore images.
The KinD ~\cite{zhang2019kindling} method proposed by Zhang \etal can be trained and achieves impressive results without the need for explicitly defining a ground truth dataset.
Liu \etal~\cite{liu2021retinex} present a retinex-based network that is both lightweight and efficient.
Guo \etal enables end-to-end training without the need for reference images~\cite{guo2020zero}.
Li \etal ~\cite{li2021learning} introduce an adaptive LLIE network that can operate under diverse lighting conditions without dependence on paired or unpaired training data.
Lim \etal~\cite{lim2020dslr} propose a method that can independently recover global illumination and local details from the original input, and gradually merge them in the image space.
Jiang ~\etal~\cite{jiang2021enlightengan} put forward unsupervised learning into the realm of LLIE using GAN.


\section{Dataset}

In our research on vignetting removal, we have encountered limitations in the existing datasets available for evaluating the performance of these methods. These datasets often consist of low-quality images or lack the necessary characteristics for effectively assessing vignetting removal algorithms. Unfortunately, at present, there is no accessible dataset exclusively designed for vignetting removal that provides reliable ground truth for objective evaluation.

Consequently, we introduce VigSet, the first high-resolution dataset that offers a comprehensive collection of paired vignetting and ground truth images, specifically designed for vignetting removal. It consists of 983 pairs of photos captured by DSLR camera and two mobile phones. What distinguishes VigSet from other vignetting datasets is its exceptional diversity, substantial quantity, and most notably, the inclusion of accurate ground truth. This ground truth information serves as a reliable reference for evaluating the effectiveness and performance of vignetting removal algorithms. Furthermore, VigSet stands out as a high-resolution dataset, with images boasting an impressive mean resolution of $5340\times3697$. This high-resolution characteristic enables researchers and practitioners to conduct detailed analyses and evaluations of vignetting removal techniques.

\subsection{Equipment for data collection}
VigSet employs a variety of three distinct capture devices: Fujifilm X-T4, ONE PLUS 10PRO, and iPhone SE. 

In well-lit situations, vignetting is generally not noticeable.
Therefore, we use an ND filter to reduce illumination and capture photos with vignetting.
The center light, which passes through an ND filter, travels a shorter distance compared to the light at the edges. This difference in distance contributes to the occurrence of vignetting.
\begin{figure}[ht]
    \begin{minipage}[b]{1.0\linewidth}
        \begin{minipage}[b]{.32\linewidth}
            \centering
            \centerline{\includegraphics[width=\linewidth,height=\linewidth]{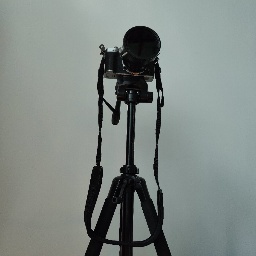}}
            \centerline{(a)}\medskip
        \end{minipage}
        \hfill
        \begin{minipage}[b]{0.32\linewidth}
            \centering
            \centerline{\includegraphics[width=\linewidth,height=\linewidth]{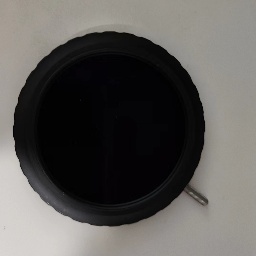}}
            \centerline{(b)}\medskip
        \end{minipage}
        \hfill
        \begin{minipage}[b]{0.32\linewidth}
            \centering
            \centerline{\includegraphics[width=\linewidth,height=\linewidth]{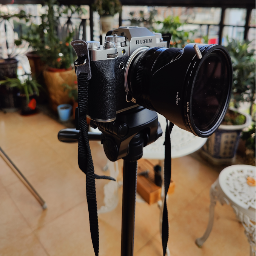}}
            \centerline{(c)}\medskip
        \end{minipage}
    \end{minipage}
    \vspace{-1.5em}
    \caption{(a) represents the visual depiction of our data acquisition equipment. (b) is the image showcasing of ND filter. and (c) illustrates the data collection process conducted by us under optimal lighting conditions along with a preview after the installation of the ND filter.}
    \label{showcase}
\end{figure}

Additionally, to ensure that there is no displacement between each pair of photos and to enhance data diversity, it is crucial to avoid dynamic objects such as swaying tree leaves, moving vehicles, and glass reflections on specific objects that are susceptible to motion.	Therefore, we select multiple distinct indoor environments for the data collection. To reduce device vibration caused by manual camera shutter presses, we use remote shutter control to capture steps~\ref{step2} and~\ref{step4}, as illustrated in Figure~\ref{showcase} (c).	
\begin{figure}[ht]
    \begin{minipage}[b]{1.0\linewidth}
    \centering 
        \begin{minipage}[b]{0.49\linewidth}
            \centering            \centerline{\includegraphics[width=\linewidth,height=\linewidth]{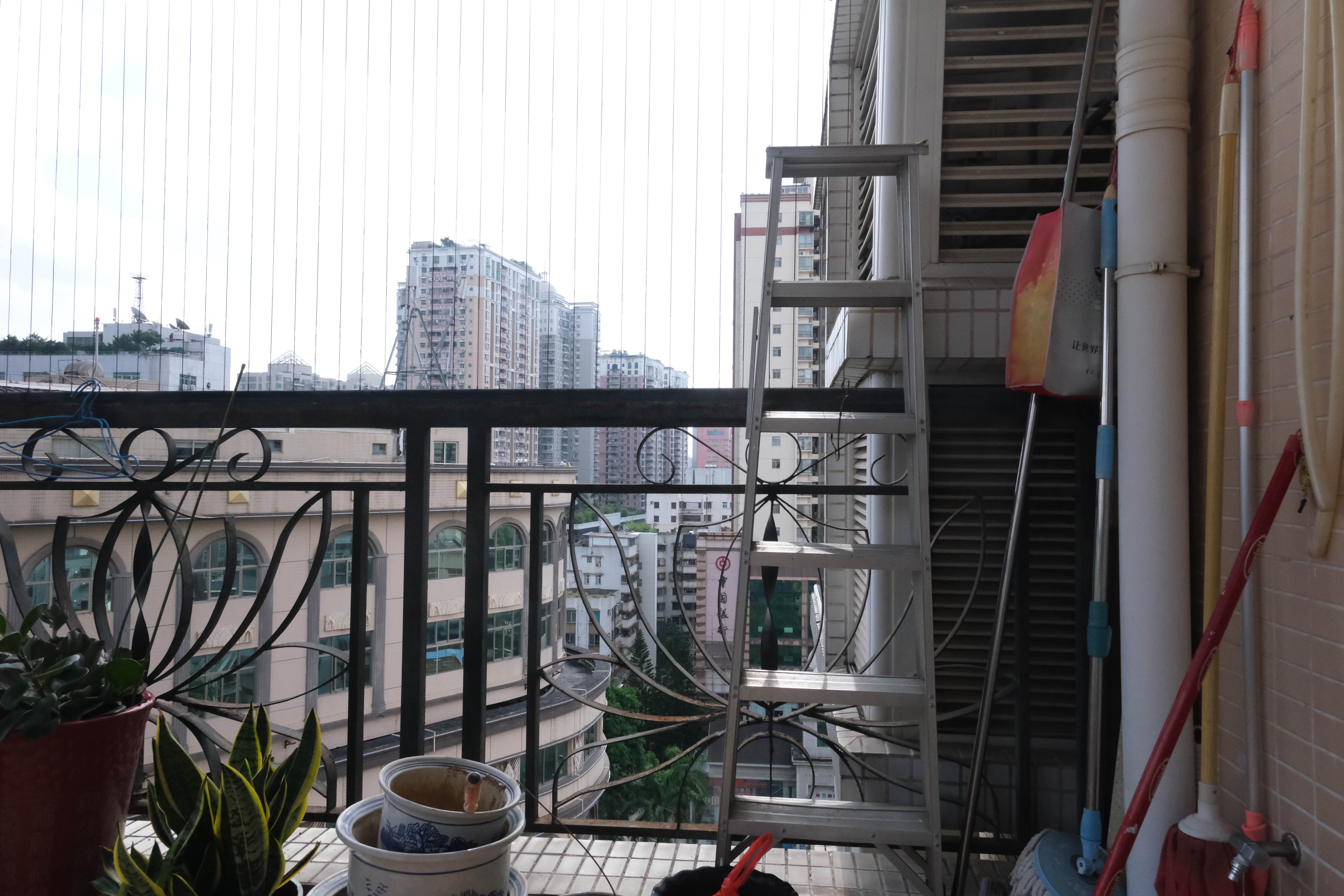}}
            \centerline{(a)}\medskip
        \end{minipage}
        \begin{minipage}[b]{0.49\linewidth}
            \centering           \centerline{\includegraphics[width=\linewidth,height=\linewidth]{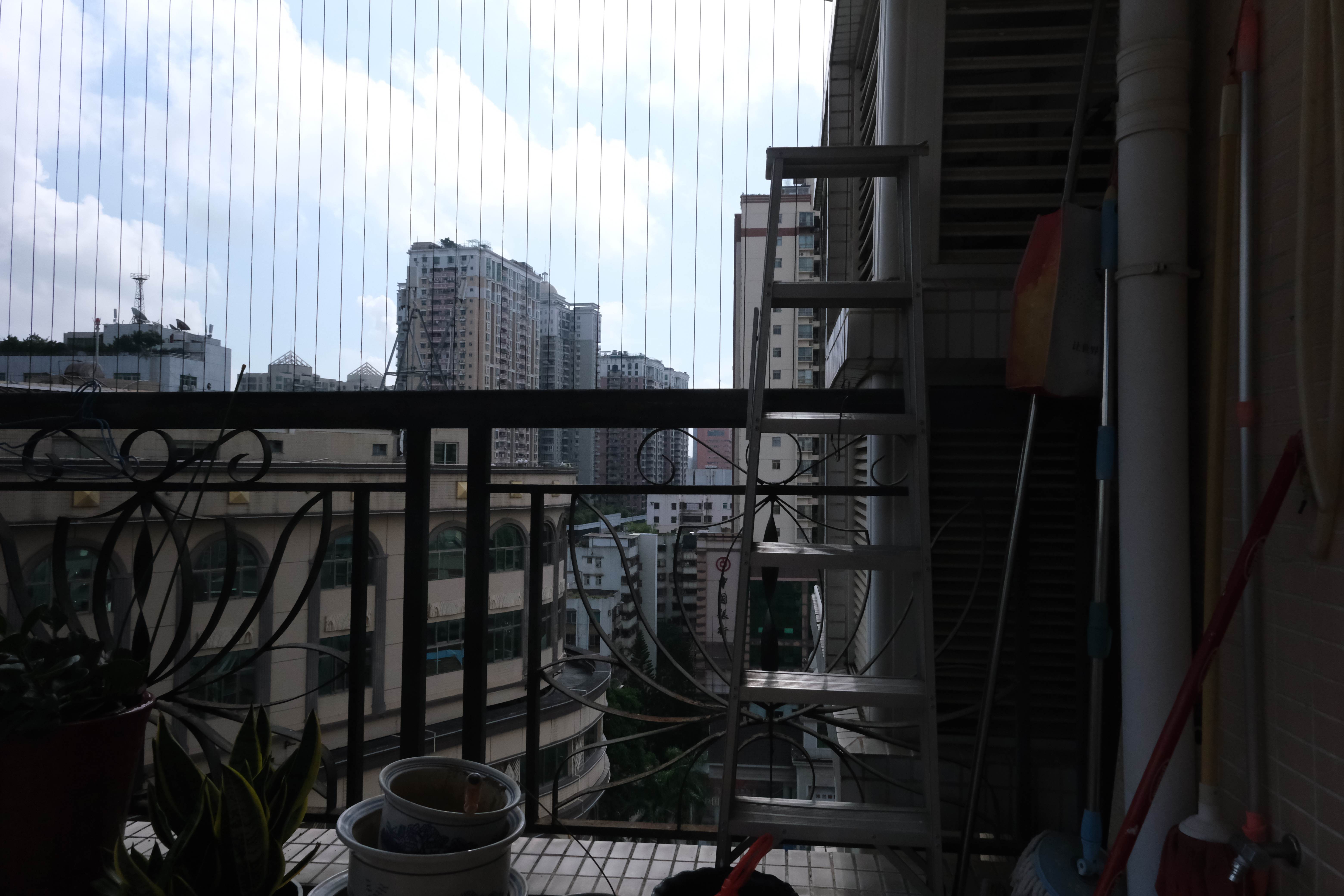}}
            \centerline{(b)}\medskip
        \end{minipage}
    \end{minipage}
    \vspace{-1.5em}
    \caption{(a) represents the image without vignetting. (b) represents the image with vignetting.}
    \label{fig:data}
\end{figure}
\subsection{Data collection and processing}
Figure~\ref{showcase} (a) provides a preview of our experimental configuration designed for capturing VigSet. The process of collecting data is facilitated by adjusting the white ND value controller located on the camera ND filter, as depicted in Figure~\ref{showcase} (c). During the data collection process, we capture images from real-world scenes utilizing experimental equipment. We use a tripod to stabilize the camera and control unrelated variables. The steps of the data collection process are as follows.
\begin{enumerate}
    \item The camera settings, such as focal length, aperture, exposure, and ISO, are set to fixed values.
    \item  A vignetting-free image is captured with an ND value of 0, as shown in Figure~\ref{fig:data} (a)\label{step2}
    \item  Continuously modify the value of the ND filter until noticeable vignetting becomes visible in the image.
    \item An image with vignetting is captured, as shown in Figure~\ref{fig:data} (b)\label{step4}
\end{enumerate}
We exclude images exhibiting obvious motion distortion or lack of focus. Moreover, we conduct group reviews of these photographs to identify and remove any outliers and duplicates.
\begin{figure*}[ht]
\begin{minipage}[b]{1.0\linewidth}
    \includegraphics[width=\linewidth]{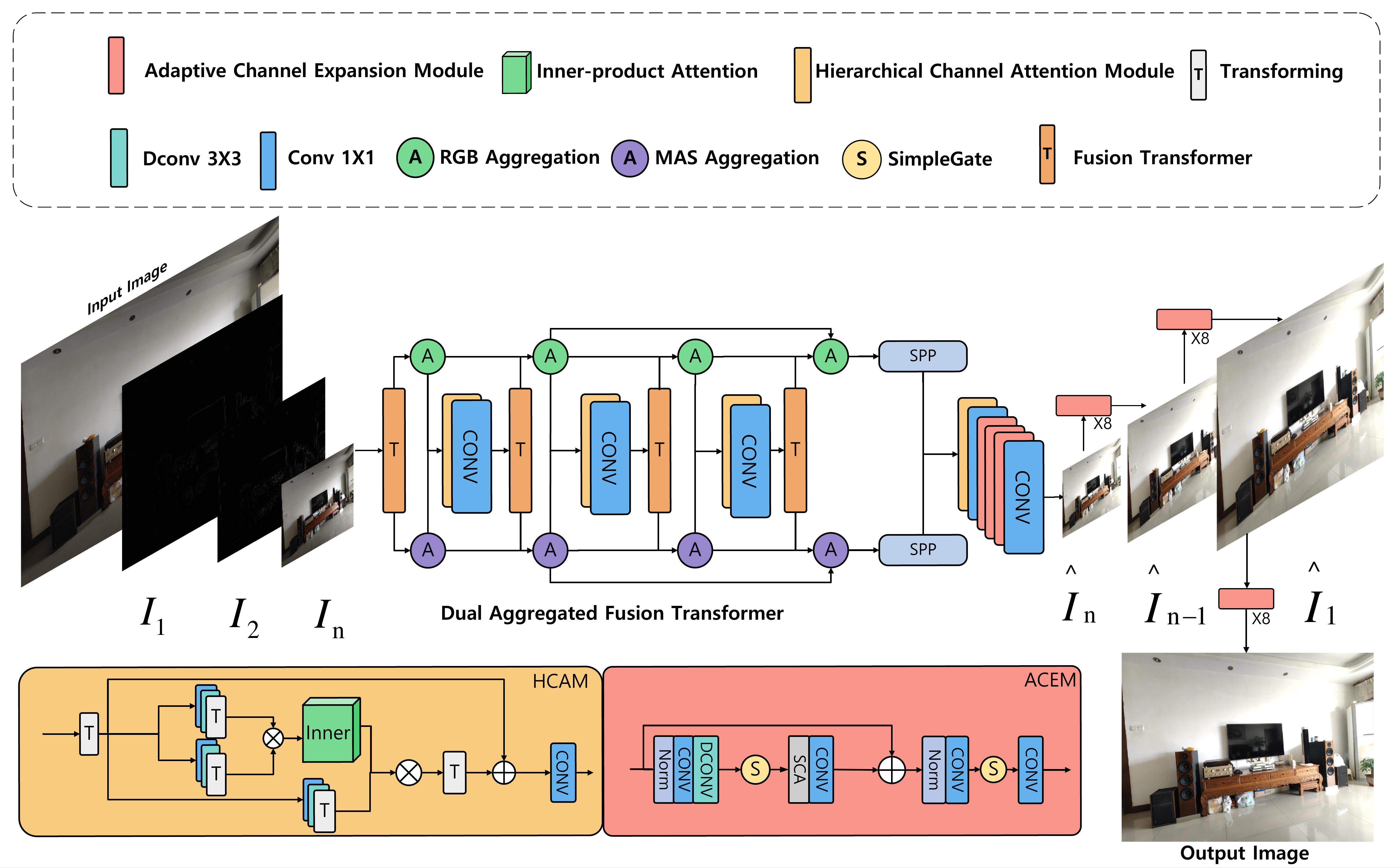}
\end{minipage}
\caption{The overall architecture of DeVigNet comprises the Dual Aggregated Fusion Transformer, Adaptive Channel Expansion Module, and Hierarchical Channel Attention Module (HCAM). Initially, the input image is decomposed into its high-frequency and low-frequency components using a Laplacian pyramid. Subsequently, the low-frequency component of the image undergoes processing with DAFT and HCAM to capture global information. Next, ACEM is employed in the reconstruction of the high-frequency component of the Laplacian pyramid to enhance edge information. Finally, a vignetting-free image result is produced.}
\label{fig:model}
\end{figure*}

\section{Methodology}

We propose a multi-frequency network based on Dual Aggregated Fusion Transformer with Adaptive Channel Expansion to exploit the features of images at various scales. Specifically, the structure of DeVigNet, as illustrated in Figure~\ref{fig:model}, includes three major components: The Dual Aggregated Fusion Transformer (DAFT), the Adaptive Channel Expansion Module (ACEM), and the Hierarchical Channel Attention Module (HCAM). In the upcoming sections, we will introduce these components of DeVigNet and the learning criteria.

\subsection{Dual Aggregated Fusion Transformer}
The Dual Aggregated Fusion Transformer (DAFT) is a neural network designed specifically for handling low-frequency information in images. As the foundational architecture, the Fusion Transformer employs multiple attention mechanisms to capture various points of focus, enabling the model to effectively incorporate both local and global information within its representation. Fusion Transformer significantly boosts the expressive capacity of the Transformer network. Figure~\ref{fig:model} illustrates the location of these four fusion transformers, wherein each transformer comprises two modules. Each module contains 1, 2, 3, or 4 Transformer Blocks~\cite{dosovitskiy2020vit} from left to right. The output of the first module is passed on to the second module, and the final output is obtained by adding the output of the second module to the output of the first module.

Inspired by~\cite{2019Towards}, the Aggregation Node is designed as integration operations on the input features. By aggregating the semantic information extracted by the fusion transformer, a richer and more comprehensive global feature representation can be obtained. This Structure facilitates an improved understanding of the image's structure, texture, and global properties by the model in low frequency, leading to enhancing vignetting removal.

\subsection{Adaptive Channel Expansion Module}
While the primary cause of vignetting is low-frequency color information, some edge information of the images remains. Therefore, motivated by U-Net~\cite{ronneberger2015u}, we propose a structure that integrates Adaptive Contrast Enhancement Module (ACEM) and Laplacian pyramid reconstruction for optimal vignetting removal results in high-frequency.

ACEM is a lightweight module that does not employ any activation functions, and it has been proven that there will be no decrease in performance. Inspired by~\cite{ba2016layer}, the beginning section of ACEM is LayerNorm, which improves stability and reduces computational overhead. The following convolutional layers capture feature information at varying scales. Inspired by~\cite{chen2022simple}, the Simplified Channel Attention (SCA) and SimpleGate techniques are utilized to enhance network performance. SimpleGate divides the feature maps into two channel dimensions and multiplies them together, leading to a reduction in computational load and complexity. The formula is as follows:
\begin{equation}\label{eqn:simple-gate}
SimpleGate(\mathbf{X},\mathbf{Y}) = \mathbf{X\odot Y},
\end{equation}
where $\mathbf{X}$ and $\mathbf{Y}$ are feature maps of the same size. 
Simplified Channel Attention (SCA) can be considered as a streamlined adaptation of Channel Attention (CA). It captures the essence of CA by retaining only two vital components: aggregating global information and enabling channel-wise interaction. SCA can be described as :
\begin{equation}\label{eqn:simplified-channel-attention}
SCA(\mathbf{X}) = \mathbf{X} * W pool(\mathbf{X}).
\end{equation}
$W$ denotes a fully connected layer. $pool$ represents the global average pooling procedure which combines spatial data into channels. $*$ indicates a channel-wise multiplication.

\begin{table*}[t]
\begin{center}
\begin{tabular}{l|c|c|c|c|c|c|c|c}
\hline
\multirow{2}{*}{Methods}  & \multicolumn{4}{c|}{\textbf{LOL}} & \multicolumn{4}{c}{\textbf{MIT-Adobe FiveK}}  \\ \cline{2-9} 

 & PSNR $\uparrow$ & SSIM $\uparrow$ & MAE $\downarrow$ & LPIPS $\downarrow$  & PSNR $\uparrow$ & SSIM $\uparrow$ & MAE $\downarrow$ & LPIPS $\downarrow$ \\ \hline
Input & 7.77 & 0.20 & 99.80 & 0.42  & 12.26 & 0.61 & 55.98 & 0.22  \\ 
LIME & 16.76 & 0.44 & 30.59 & 0.31  & 13.30 & 0.75 & 52.12 & 0.09  \\ 
MSRCR  &13.17 & 0.45 & 52.71 & 0.33  & 13.31 & 0.75 & 50.82 & 0.12  \\ 
NPE & 16.97 & 0.47 & 32.89 & 0.31  & 17.38 & 0.79 & 31.22 & 0.10  \\ 
WV-SRIE & 11.86 & 0.49 & 65.56 & 0.24  & 18.63 & 0.84 & 26.27 & 0.08  \\ 
PM-SRIE & 12.28 & 0.51 & 63.28 & 0.23  & 19.70 & 0.84 & 23.42 & 0.07  \\ 
JieP & 12.05 & 0.51 & 64.34 & 0.22  & 18.64 & 0.84 & 26.42 & \underline{0.07}  \\ 
RetinexNet & 16.77 & 0.42 & 32.02 & 0.38  & 12.51 & 0.69 & 52.73 & 0.20  \\ 
KID & 17.65 & \textbf{0.77} & 31.40 & \textbf{0.12}  & 16.20 & 0.79 & 35.16 & 0.11  \\ 
DSLR  & 14.98 & 0.60 & 48.90 & 0.27  & 20.24 & 0.83 & 22.45 & 0.10  \\ 
ELGAN & 17.48 & 0.65 & 34.47 & 0.23  & 16.00 & 0.79 & 36.37 & 0.09  \\ 
RUAS  & 16.40 & 0.50 & 39.11 & 0.19 & 17.91 & 0.84 & 33.12 & 0.08  \\ 
Zero\_DCE & 14.86 & 0.56 & 47.07 & 0.24  & 15.93 & 0.77 & 36.36 & 0.12  \\
Zero\_DCE++& 14.75 & 0.52 & 45.94 & 0.22  & 14.61 & 0.42 & 39.25 & 0.16  \\ 
Uformer & \underline{18.55} & 0.73 & \underline{28.91} & 0.23  & \underline{21.92} & \textbf{0.87} & \underline{17.91} & \textbf{0.06}  \\ 
\textbf{Ours} & \textbf{21.33} & \underline{0.76} & \textbf{19.30} & \underline{0.16}  & \textbf{23.10} & \underline{0.84} & \textbf{15.43} & 0.16  \\ 
\hline
\end{tabular}
\caption{Quantitative results on LOL and MIT-Adobe FiveK datasets in terms of PSNR, SSIM, LPIPS and MAE. The top two results are marked in bold and underline.}
\label{quan}
\end{center}
\end{table*}

\subsection{Hierarchical Channel Attention Module}
Inspired by~\cite{wang2023ultra}, The Hierarchical Channel Attention Module (HCAM) module is dedicated to the hierarchical fusion of features and the acquisition of learnable correlations across diverse layers. The primary function of HCAM is to compute and apply attention weights to the input feature map, resulting in the refinement and enhancement of vignetting features. Initially, HCAM performs a transformation on the input $I\in R^{H\times W\times 3}$ to yield $R^{N\times H\times W\times C}, N = 3$, from successive layers. Subsequently, the $Q$ (query), $K$ (key), and $V$ (value) features are extracted using convolutional layers. Subsequently, the $Q$, $K$, and $V$ features are extracted using convolutional layers. 

In this module, Inner-Product Attention plays a crucial role in calculating attention scores by computing the dot product between the query and the key. HCAM adaptively fuses features from different hierarchical levels by conducting weighted operations between values and attention scores. The corresponding output feature ${\mathbf{R}}_{\text {out }}$ can be described as:
\begin{equation}
    {\mathbf{R}}_{\text {out }}=W_{1 \times 1}LA
({\mathbf{Q}}, {\mathbf{K}}, {\mathbf{V}})+{\mathbf{R}}_{\text {in }},
\label{eq:fout}
\end{equation}
where $LA$ represents Layer Attention, $W_{1 \times 1}$ indicates Convolution $1 \times 1$, $LA$ can be written as:
\begin{equation}
    LA(\hat{\mathbf{Q}}, \hat{\mathbf{K}}, \hat{\mathbf{V}})=\hat{\mathbf{V}} \operatorname{softmax}(\hat{\mathbf{Q}} \hat{\mathbf{K}} / \boldsymbol{\alpha}),
\label{eq:la}
\end{equation}

\subsection{Learning Criteria}

During the training phase, we employ two loss functions, namely Mean Squared Error ($L_{MSE}$) and Structural Similarity Index Measure ($L_{SSIM}$)~\cite{wang2004image}, which offer significant advantages in preserving the intricate details of the image. The mathematical expressions for these losses are as follows:
\begin{equation}
L_{total} = L_{MSE} + \lambda * L_{SSIM},
\label{eq:total}
\end{equation}
The weight of $L_{SSIM}$ is empirically set to 0.4.

\section{Experiments}

\begin{table*}[!ht]
\begin{center}
\adjustbox{width=\linewidth}{
\begin{tabular}{l|c|c|c|c|c|c|c|c|c|c|c|c}
\hline
\multirow{2}{*}{Methods}  & \multicolumn{4}{c|}{\textbf{$512\times512$}} & \multicolumn{4}{c|}{\textbf{$1024\times1024$}}& \multicolumn{4}{c}{\textbf{$2048\times2048$}}  \\ \cline{2-13} 

 & PSNR $\uparrow$ & SSIM $\uparrow$ & MAE $\downarrow$ & LPIPS $\downarrow$  & PSNR $\uparrow$ & SSIM $\uparrow$ & MAE $\downarrow$ & LPIPS $\downarrow$& PSNR $\uparrow$ & SSIM $\uparrow$ & MAE $\downarrow$ & LPIPS $\downarrow$ \\ \hline
Input & 12.08 & 0.59 & 60.04 & 0.18  & 12.08 & 0.58 & 60.04 & 0.18  & 12.08 & 0.58 & 60.04 & 0.19  \\ 
RIVC & 13.08 & 0.59 & 55.17 & 0.18  & 13.08 & 0.59 & 55.17 & 0.18  & 13.08 & 0.59 & 55.17 & 0.19  \\
SIVC & 14.65 & 0.62 & 43.71 & 0.17  & 14.65 & 0.61 & 43.69 & \underline{0.17}  & 14.65 & 0.60 & 43.70 & 0.18  \\ 
LIME & 12.42 & 0.41 & 51.29 & 0.41  & 12.42 & 0.39 & 51.29 & 0.40  & 12.41 & 0.37 & 51.32 & 0.42  \\ 
MSRCR & 11.20 & 0.39 & 60.43 & 0.44  & 11.20 & 0.37 & 60.43 & 0.45  & 11.20 & 0.35 & 60.44 & 0.46  \\
NPE & 15.72 & 0.51 & 38.60 & 0.33  & 15.72 & 0.49 & 38.59 & 0.32  & 15.72 & 0.47 & 38.60 & 0.33  \\ 
WV-SRIE & 18.84 & 0.60 & 26.67 & 0.22  & 18.84 & 0.58 & 26.67 & 0.23  & 18.84 & 0.56 & 26.67 & 0.24  \\
PM-SRIE & 19.45 & 0.66 & 24.81 & 0.16  & \underline{19.45} & 0.64 & 24.81 & \underline{0.17}  & \underline{19.45} & 0.62 & 24.81 & 0.19  \\ 
JieP & 18.93 & 0.58 & 26.33 & 0.24  & 18.93 & 0.55 & 26.32 & 0.25  & 18.93 & 0.54 & 26.32 & 0.27  \\  
KID & 14.73 & 0.71 & 44.01 & 0.18  & 14.74 & 0.71 & 43.95 & 0.22  & 14.73 & \underline{0.71} & 44.00 & 0.31  \\
DSLR & 19.37 & 0.65 & 24.10 & 0.16  & 19.37 & 0.64 & \underline{24.07} & 0.20  & 19.35 & 0.62 & \underline{24.10} & 0.29  \\ 
ELGAN & 16.32 & 0.73 & 37.77 & \underline{0.10}  & 16.32 & \underline{0.72} & 37.76 & \textbf{0.11}  & 16.31 & 0.72 & 37.77 & \textbf{0.12}  \\  
RUAS & 15.54 & 0.60 & 36.93 & 0.22  & 15.54 & 0.57 & 36.92 & 0.24  & 15.54 & 0.56 & 36.92 & 0.25  \\ 
Zero-DCE & 16.28 & 0.58 & 34.77 & 0.26  & 16.28 & 0.57 & 34.77 & 0.26  & 16.28 & 0.55 & 34.78 & 0.26  \\ 
Zero-DCE++ & 16.82 & 0.55 & 32.31 & 0.20  & 16.82 & 0.52 & 32.32 & 0.21  & 16.81 & 0.51 & 32.34 & 0.23  \\ 
Uformer & \underline{20.95} & \underline{0.77} & \underline{21.32} & 0.19  & 20.60 & 0.77 & 22.80 & 0.25  & 20.67 & 0.77 & 22.69 & 0.28  \\ 
\textbf{Ours} & \textbf{22.96} & \textbf{0.79} & \textbf{15.82} & \textbf{0.09}  & \textbf{22.94} & \textbf{0.78} & \textbf{15.84} & \textbf{0.11}  & \textbf{22.94} & \textbf{0.77} & \textbf{15.85} & \underline{0.13}  \\ 
\hline
\end{tabular}}
\caption{Quantitative results on VigSet in terms of PSNR, SSIM, LPIPS and MAE. The top two results are marked in bold and underline.}
\label{tb:vigset}
\end{center}
\end{table*}

\begin{table*}[t]
\adjustbox{width=\linewidth}{
\begin{tabular}{l|c|c|c|c|c|c|c|c|c|c|c|c}
\hline
\multirow{2}{*}{Ablation Study}  & \multicolumn{4}{c|}{\textbf{$512\times512$}} & \multicolumn{4}{c|}{\textbf{$1024\times1024$}}& \multicolumn{4}{c}{\textbf{$2048\times2048$}}  \\ \cline{2-13} 
& PSNR $\uparrow$ & SSIM $\uparrow$ & MAE $\downarrow$ & LPIPS $\downarrow$  & PSNR $\uparrow$ & SSIM $\uparrow$ & MAE $\downarrow$ & LPIPS $\downarrow$& PSNR $\uparrow$ & SSIM $\uparrow$ & MAE $\downarrow$ & LPIPS $\downarrow$ \\ \hline
Input & 12.08 & 0.59 & 60.04 & 0.18  & 12.08 & 0.58 & 60.04 & 0.18  & 12.08 & 0.58 & 60.04 & 0.19  \\ 
Ours w/ Depth=3 & 16.04 & 0.69 & 32.44 & 0.16 & 16.04 & 0.69 & 32.44 & 0.17 & 16.04 & 0.70 & 32.44 & 0.19  \\ 
Ours w/ Depth=4 & 13.94 & 0.66 & 42.62 & 0.21 & 13.94 & 0.67 & 42.63 & 0.23 & 13.94 & 0.68 & 42.63 & 0.22  \\ 
Ours w/o ACEM & 19.66 & 0.76 & 23.77 & 0.11 & 19.66 & 0.74 & 23.78 & 0.14 & 19.65 & 0.74 & 23.79 & 0.17  \\ 
Ours w/o DAFT & 18.25 & 0.73 & 28.75 & 0.14 & 18.25 & 0.72 & 28.75 & 0.15 & 18.25 & 0.72 & 28.75 & 0.16  \\ 
DHAN w/ ACEM & 21.84 & 0.75 & 17.73& 0.20 & 20.95 & 0.74 & 19.96 & 0.21 & 21.13 & 0.74 & 19.25 & 0.23 \\
\textbf{Ours} & \textbf{22.96} & \textbf{0.79} & \textbf{15.82} & \textbf{0.09}  & \textbf{22.94} & \textbf{0.78} & \textbf{15.84} & \textbf{0.11}  & \textbf{22.94} & \textbf{0.77} & \textbf{15.85} & \textbf{0.13}  \\ \hline
\end{tabular}}
\caption{Quantitative results of the ablation study on three datasets in different resolutions. The top two results are marked in bold and underline.}
\label{ablation}
\end{table*}

\begin{figure*}[!ht] 
    \begin{minipage}[b]{1.0\linewidth}
        \begin{minipage}[b]{0.24\linewidth}
            \centering
            \centerline{\includegraphics[width=\linewidth]{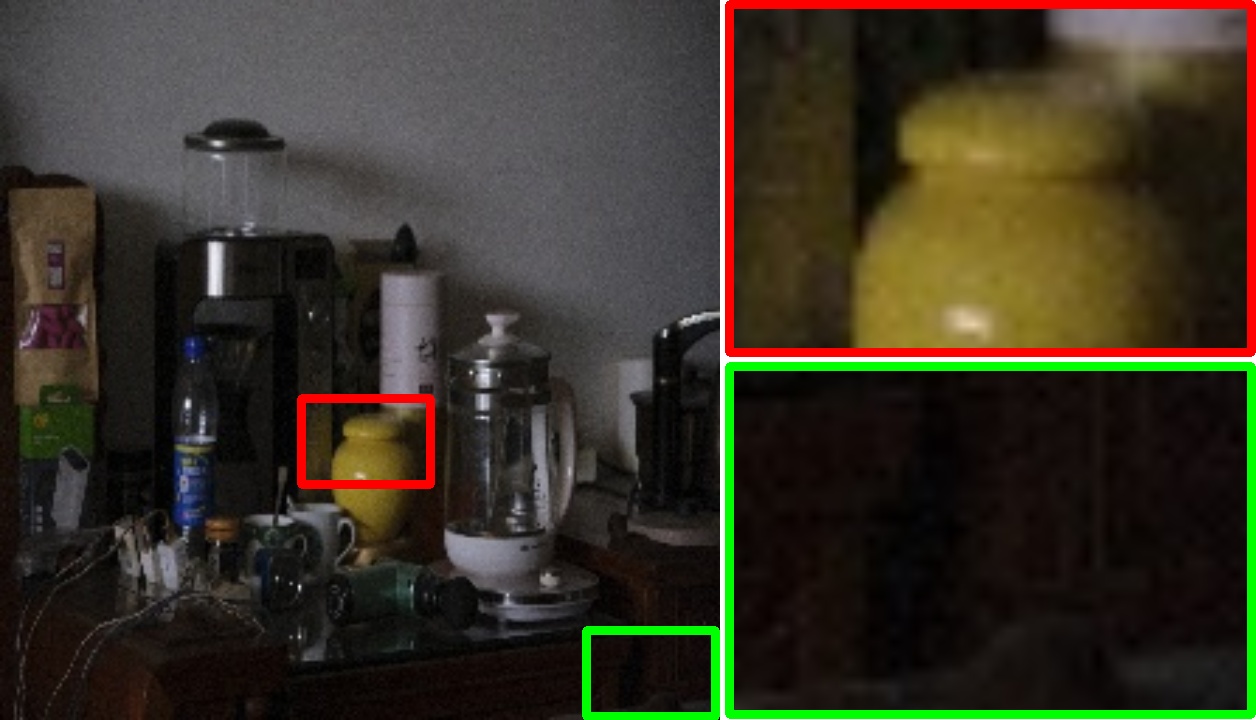}}
            \centerline{(a) Input}\medskip
        \end{minipage}
        \hfill
        \begin{minipage}[b]{0.24\linewidth}
            \centering
            \centerline{\includegraphics[width=\linewidth]{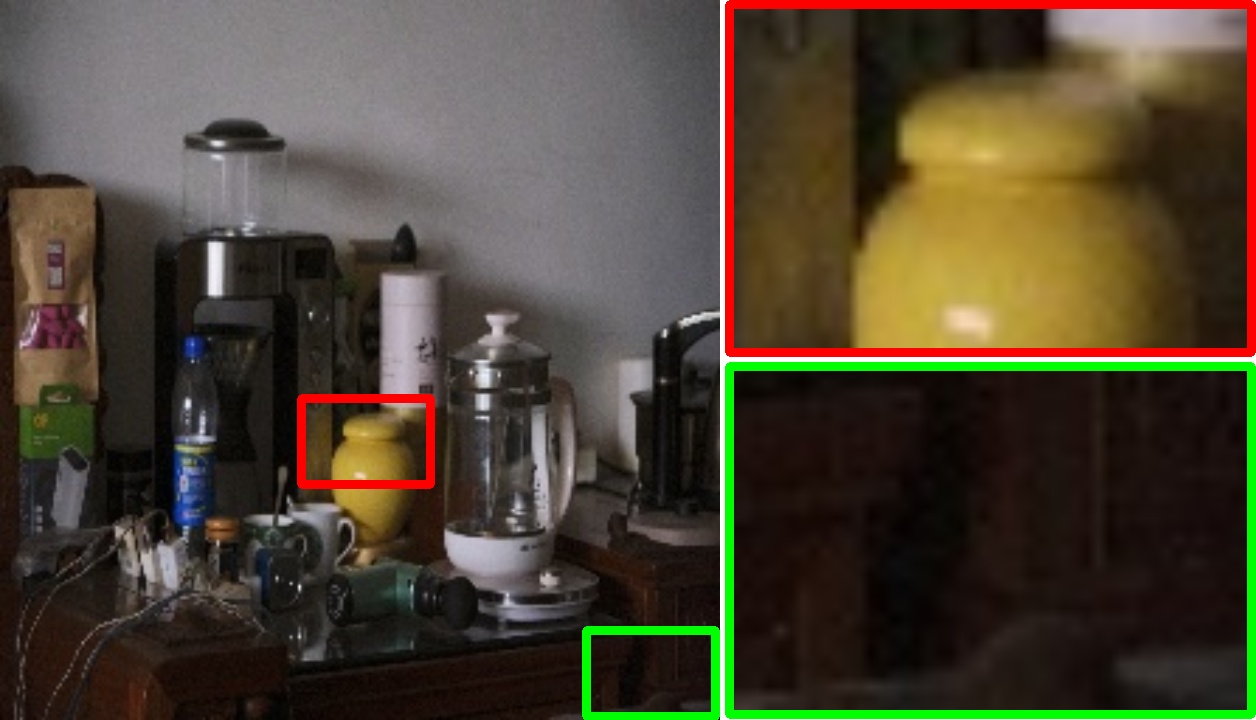}}
            \centerline{(b) ELGAN}\medskip
        \end{minipage} 
        \hfill
        \begin{minipage}[b]{0.24\linewidth}
            \centering
            \centerline{\includegraphics[width=\linewidth]{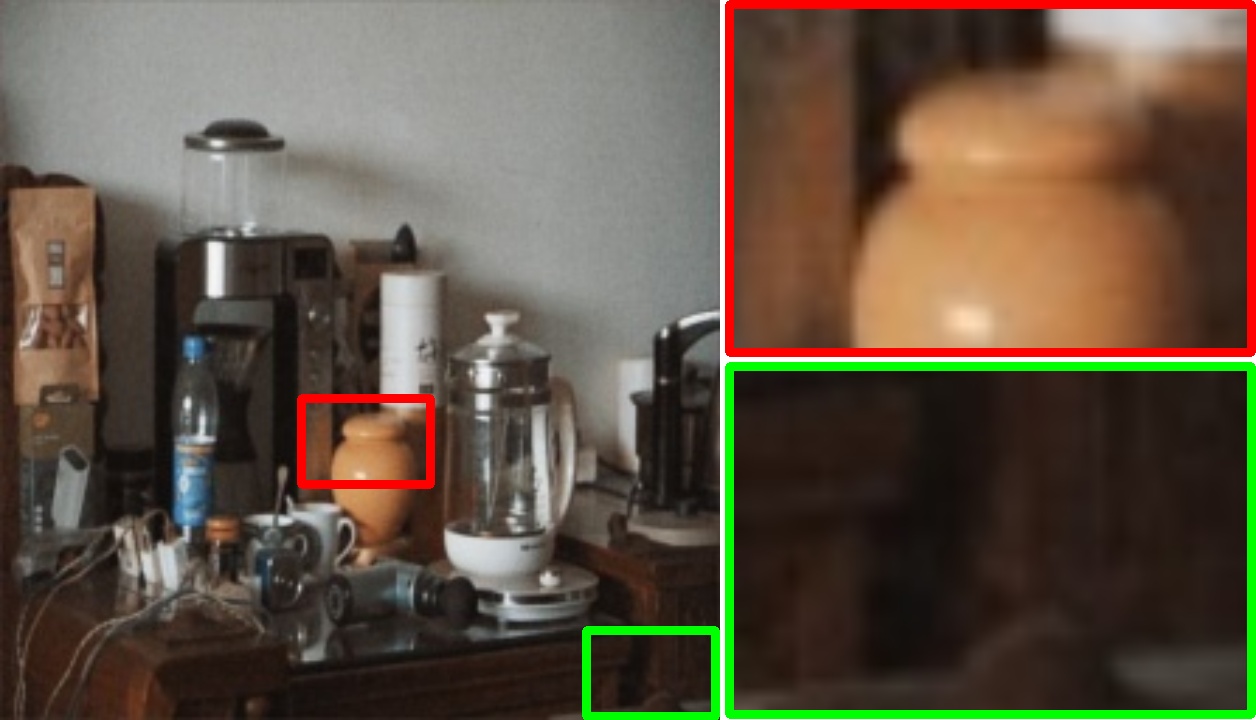}}
            \centerline{(c) UFormer }\medskip
        \end{minipage}  
        \hfill
        \begin{minipage}[b]{0.24\linewidth}
            \centering
            \centerline{\includegraphics[width=\linewidth]{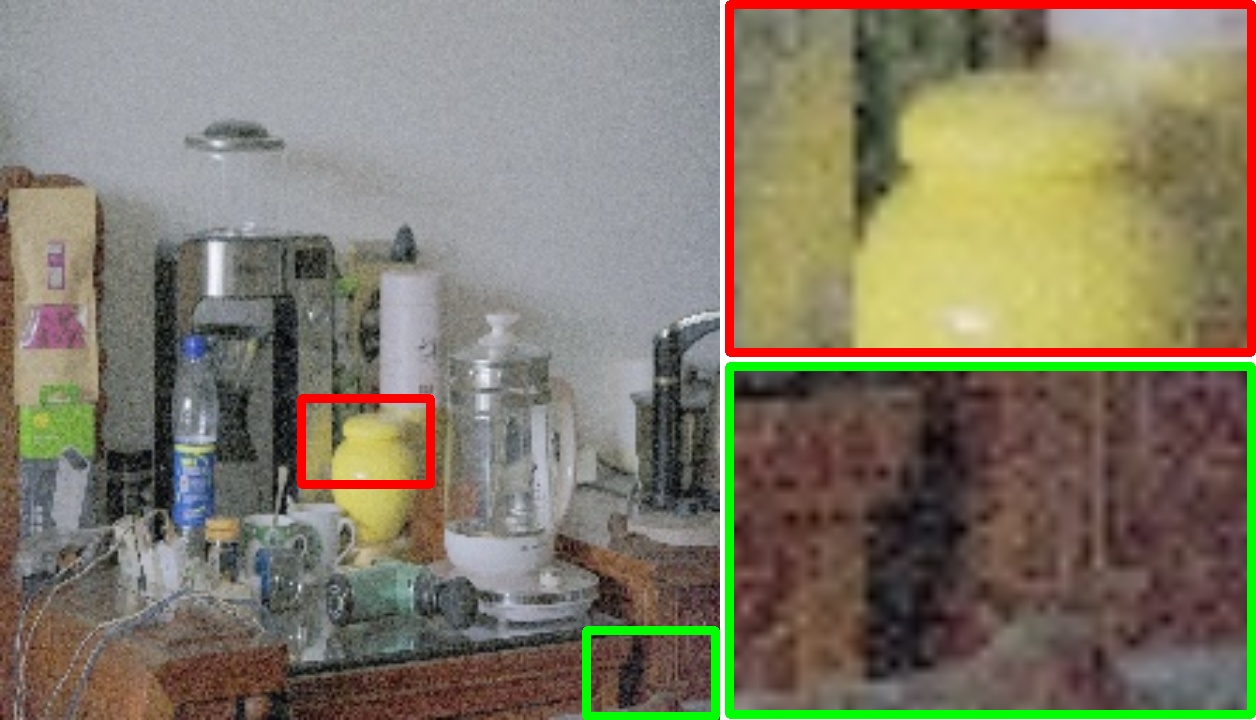}}
            \centerline{(d) Zero-DCE}\medskip
        \end{minipage}  
    \end{minipage}
    \begin{minipage}[b]{1.0\linewidth}
        \begin{minipage}[b]{0.24\linewidth}
            \centering
            \centerline{\includegraphics[width=\linewidth]{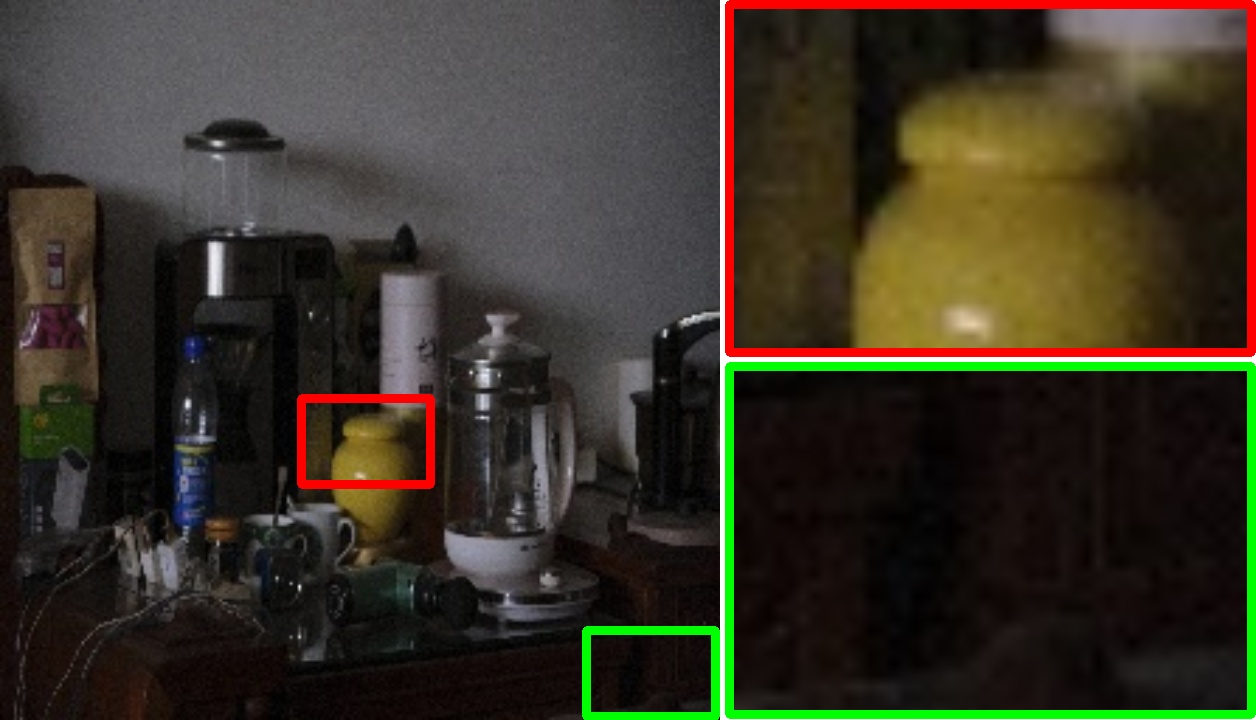}}
            \centerline{(e) RIVC}\medskip
        \end{minipage}
        \hfill
        \begin{minipage}[b]{0.24\linewidth}
            \centering
            \centerline{\includegraphics[width=\linewidth]{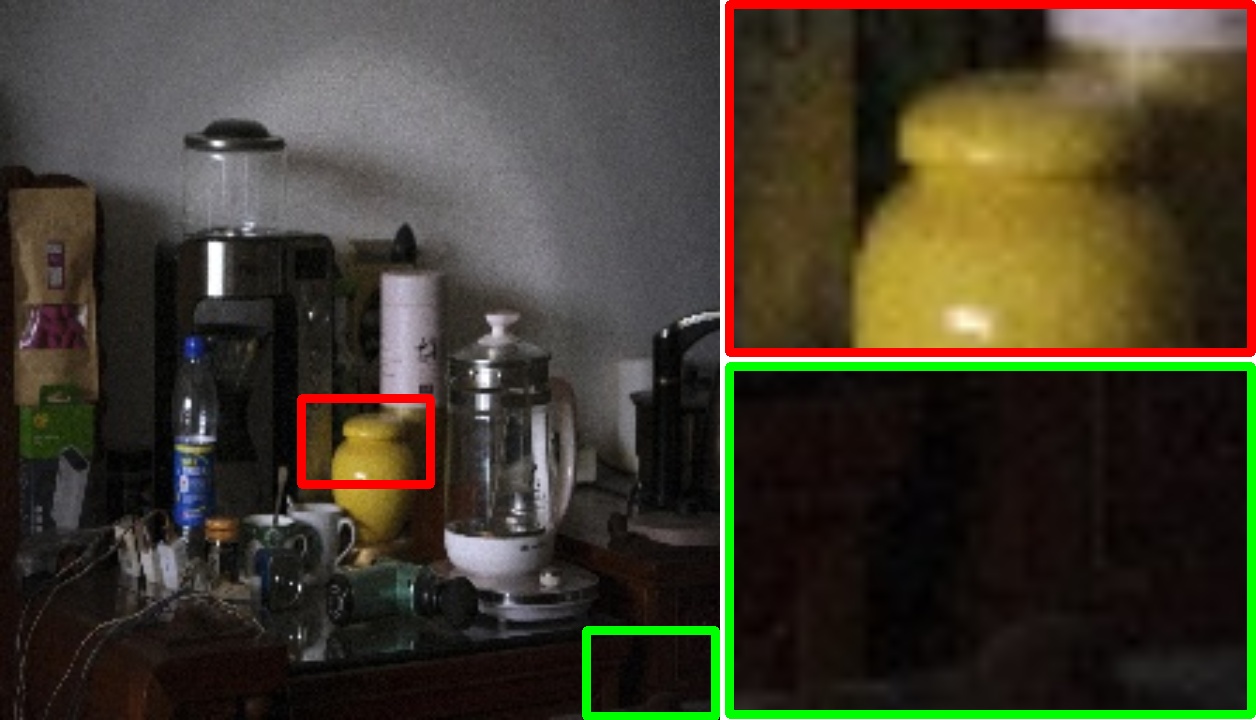}}
            \centerline{(f) SIVC}\medskip
        \end{minipage}    
        \hfill
        \begin{minipage}[b]{0.24\linewidth}
            \centering
            \centerline{\includegraphics[width=\linewidth]{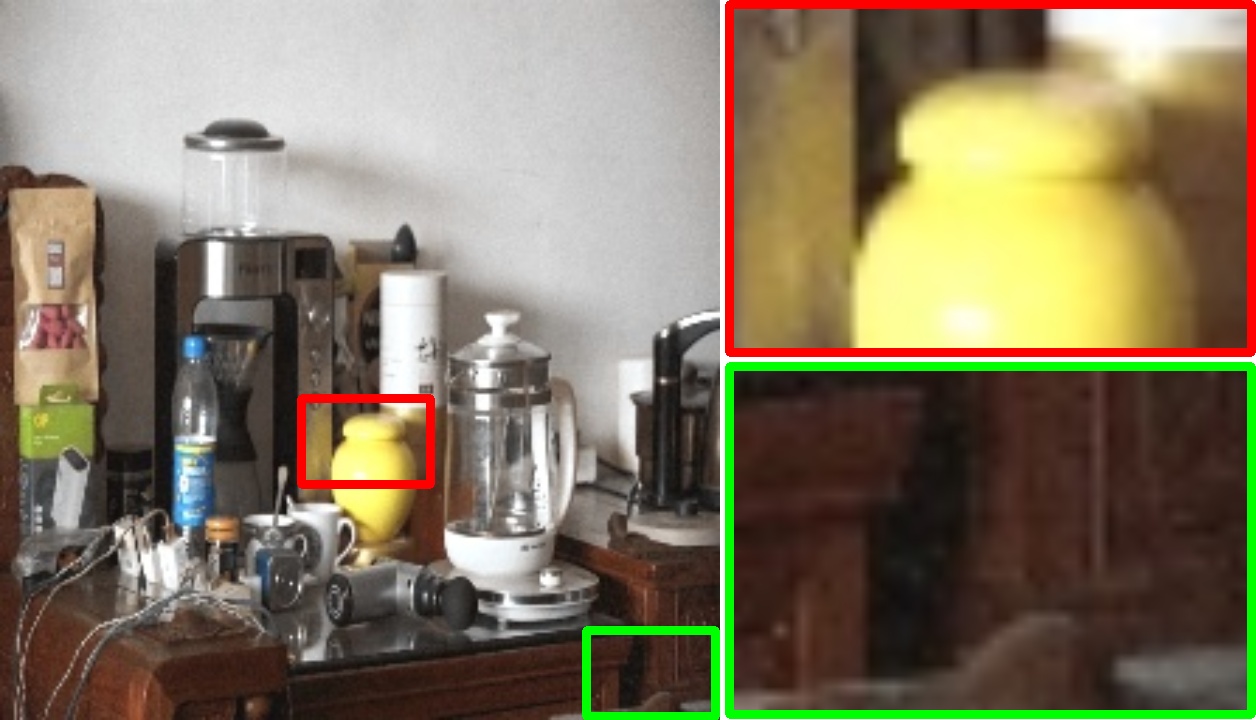}}
        \centerline{(g) Ours}\medskip
        \end{minipage}
        \hfill
        \begin{minipage}[b]{0.24\linewidth}
            \centering
            \centerline{\includegraphics[width=\linewidth]{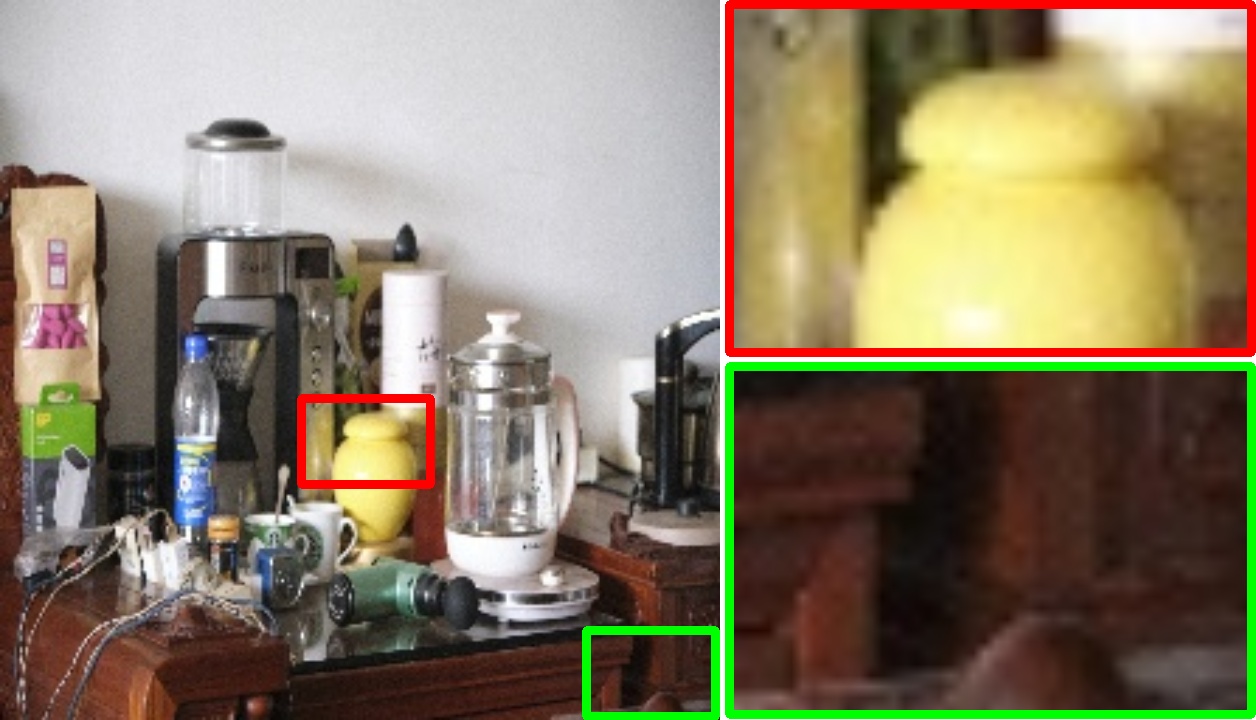}}
            \centerline{(h) Target}\medskip
        \end{minipage}
    \end{minipage}

    \caption{
   Visual comparison of various Vignetting Removal and LLIE methods on the VigSet dataset is presented. The figure clearly illustrates the presence of noticeable vignetting in methods (b), (f), and (e). Color degradation or distortion issues are apparent in methods (c) and (d). 
   }
    \label{fig:vigset}
\end{figure*}

\begin{figure*}[!ht] 

    \begin{minipage}[b]{1.0\linewidth}
        \begin{minipage}[b]{0.24\linewidth}
            \centering
            \centerline{\includegraphics[width=\linewidth]{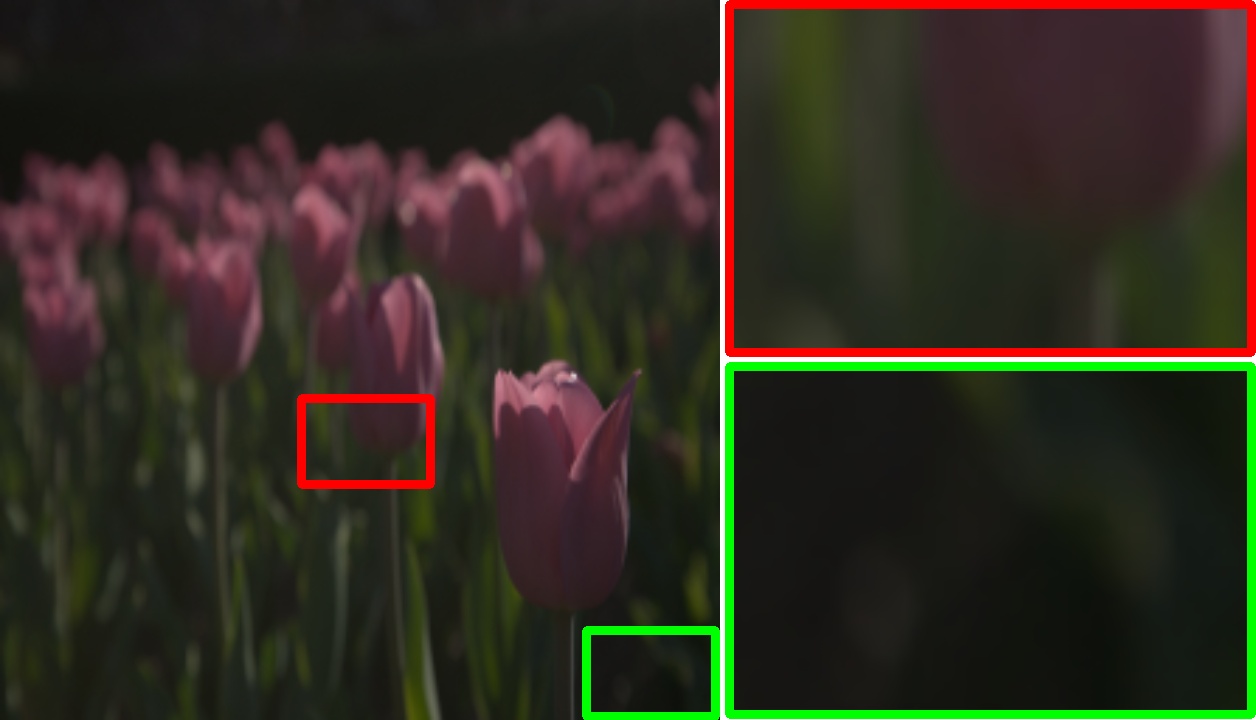}}
            \centerline{(a) Input}\medskip
        \end{minipage}
        \hfill
        \begin{minipage}[b]{0.24\linewidth}
            \centering
            \centerline{\includegraphics[width=\linewidth]{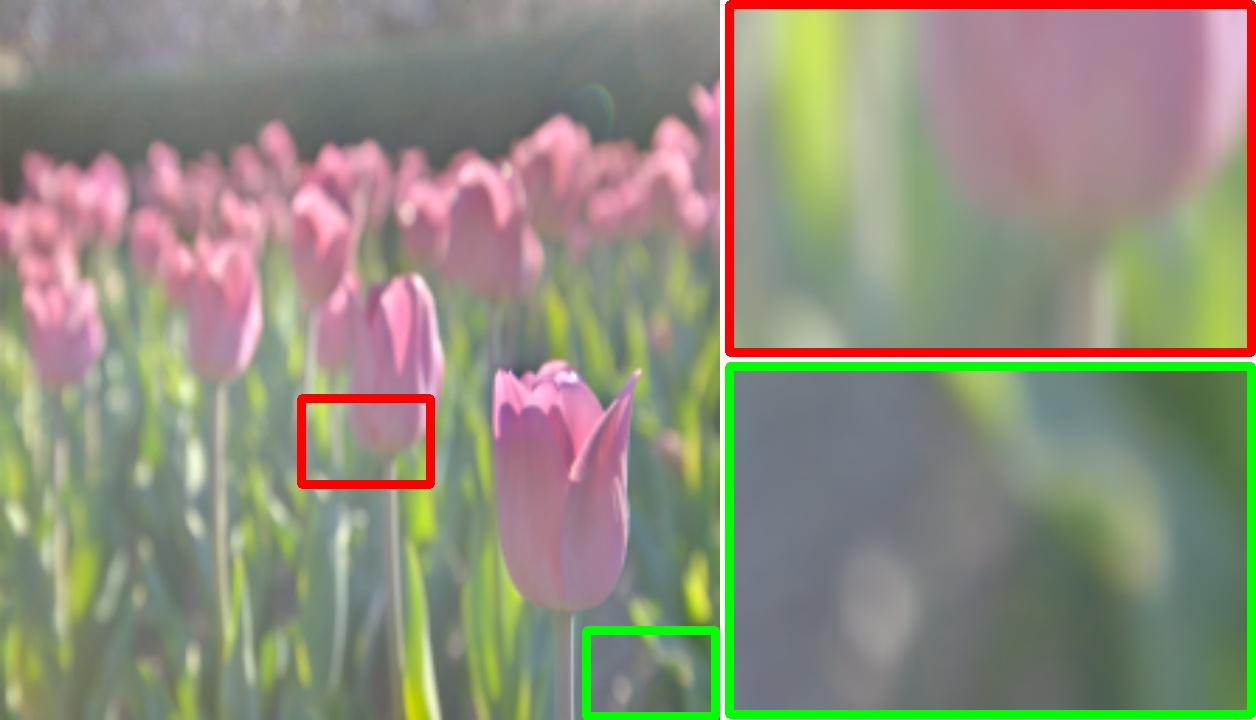}}
            \centerline{(b) ELGAN}\medskip
        \end{minipage} 
        \hfill
        \begin{minipage}[b]{0.24\linewidth}
            \centering
            \centerline{\includegraphics[width=\linewidth]{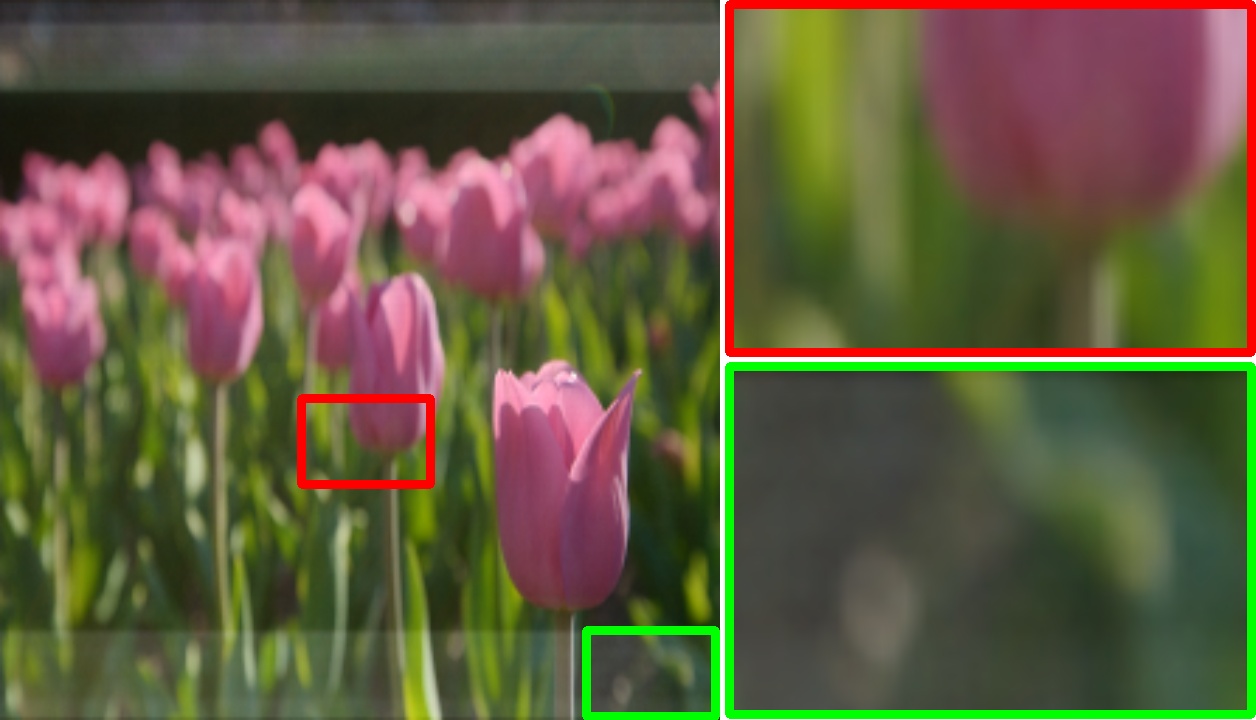}}
            \centerline{(c) UFormer }\medskip
        \end{minipage}  
        \hfill
        \begin{minipage}[b]{0.24\linewidth}
            \centering
            \centerline{\includegraphics[width=\linewidth]{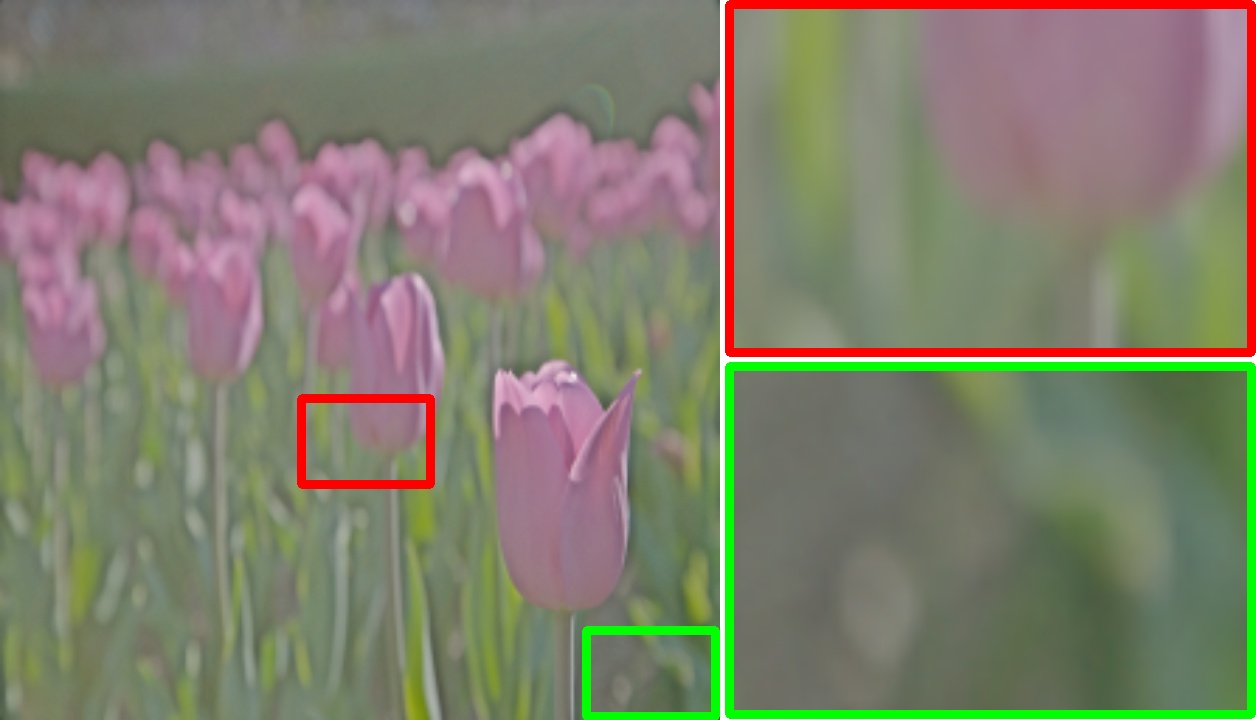}}
            \centerline{(d) Zero-DCE}\medskip
        \end{minipage}  
    \end{minipage}
    \begin{minipage}[b]{1.0\linewidth}
        \begin{minipage}[b]{0.24\linewidth}
            \centering
            \centerline{\includegraphics[width=\linewidth]{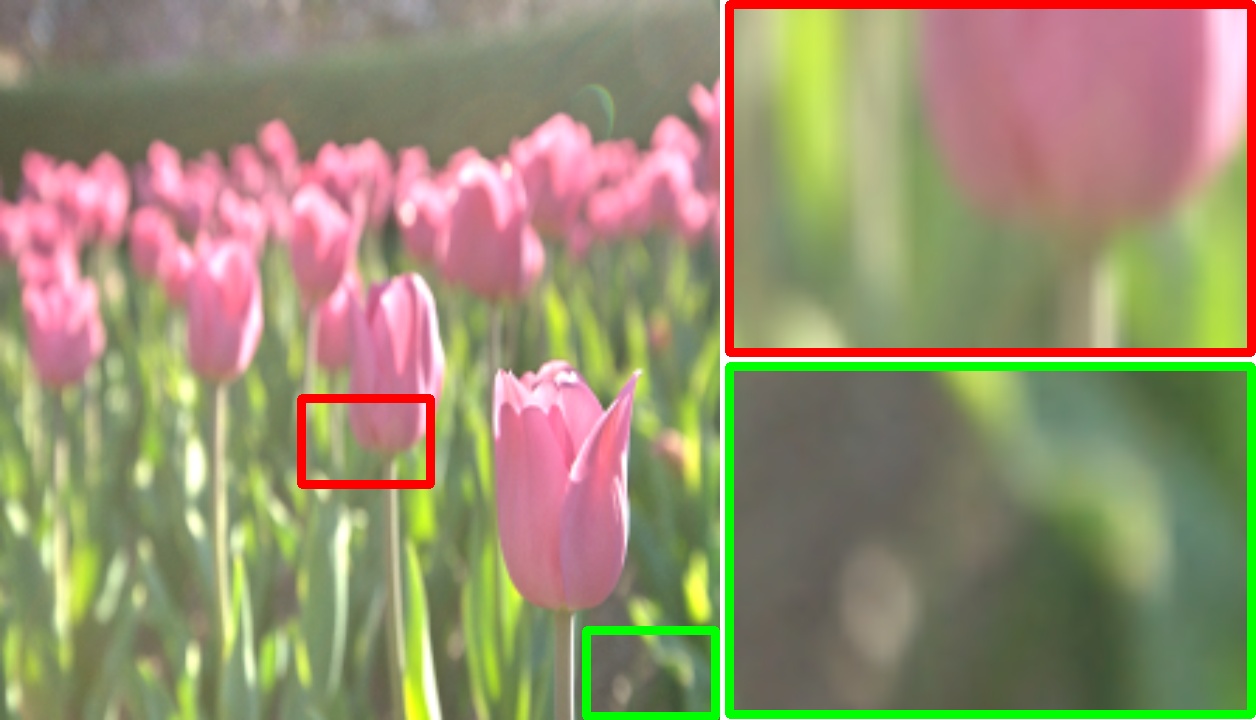}}
            \centerline{(e) LIME}\medskip
        \end{minipage}
        \hfill
        \begin{minipage}[b]{0.24\linewidth}
            \centering
            \centerline{\includegraphics[width=\linewidth]{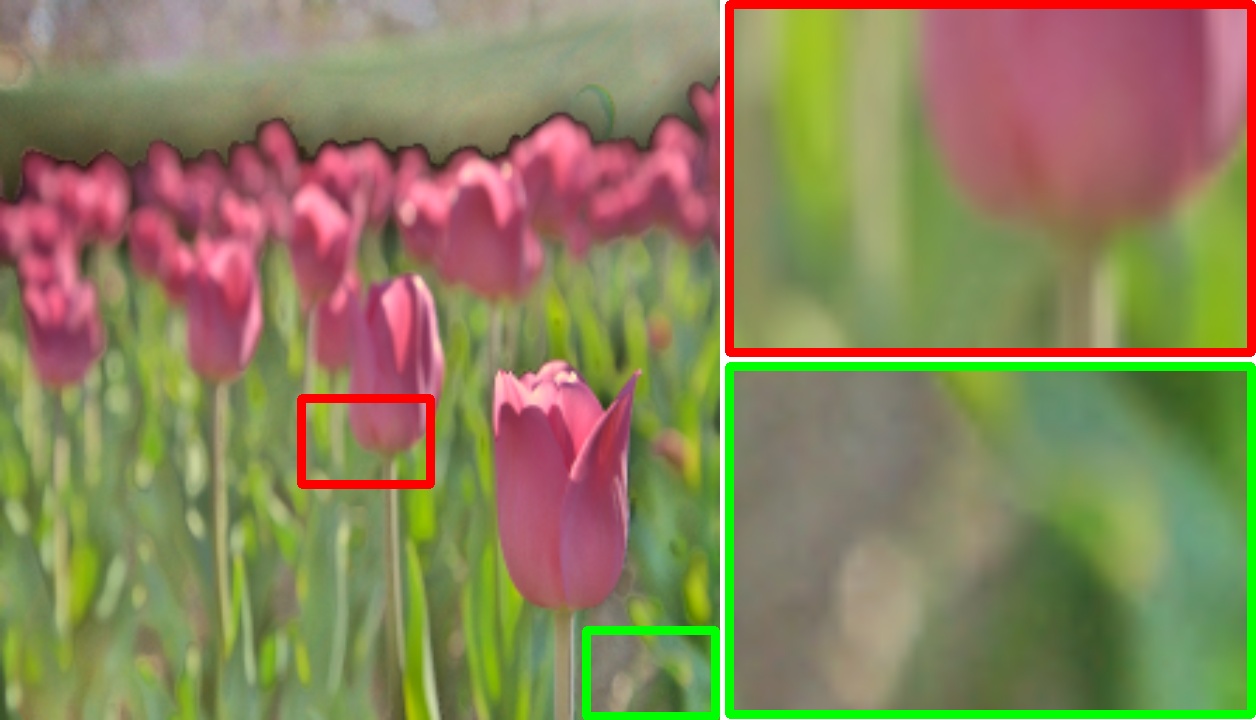}}
            \centerline{(f) NPE}\medskip
        \end{minipage}    
        \hfill
        \begin{minipage}[b]{0.24\linewidth}
            \centering
            \centerline{\includegraphics[width=\linewidth]{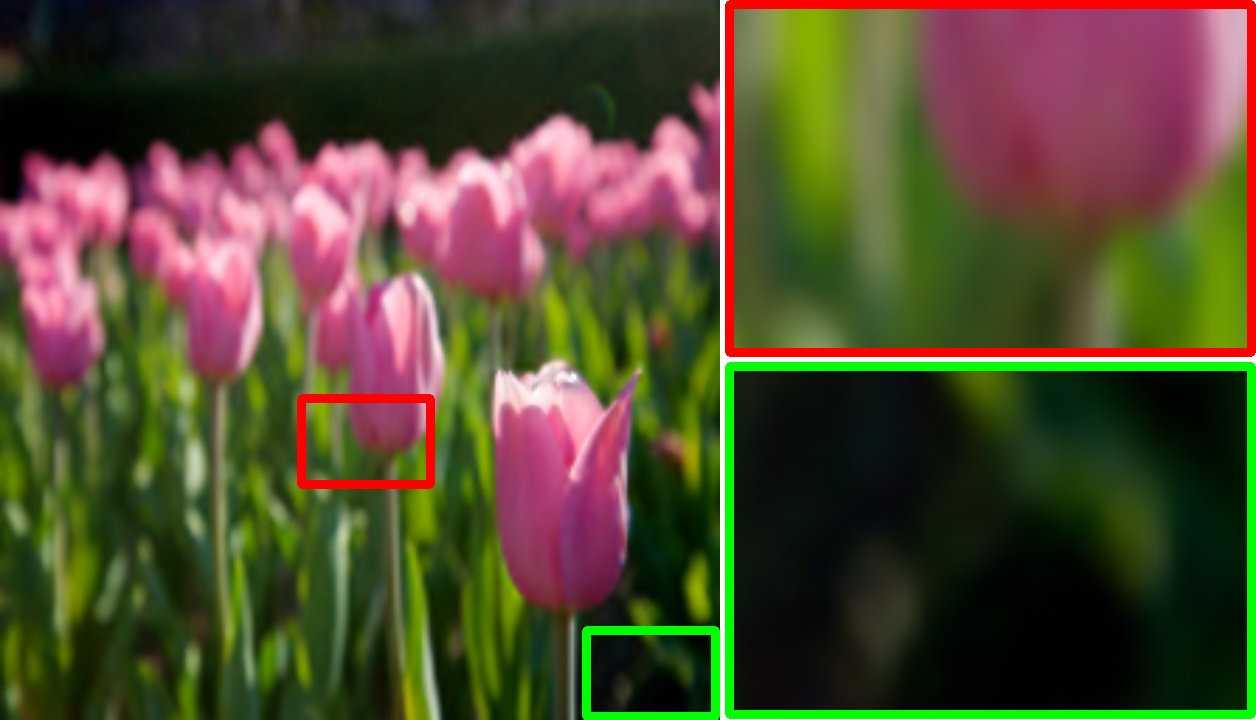}}
        \centerline{(g) Ours}\medskip
        \end{minipage}
        \hfill
        \begin{minipage}[b]{0.24\linewidth}
            \centering
            \centerline{\includegraphics[width=\linewidth]{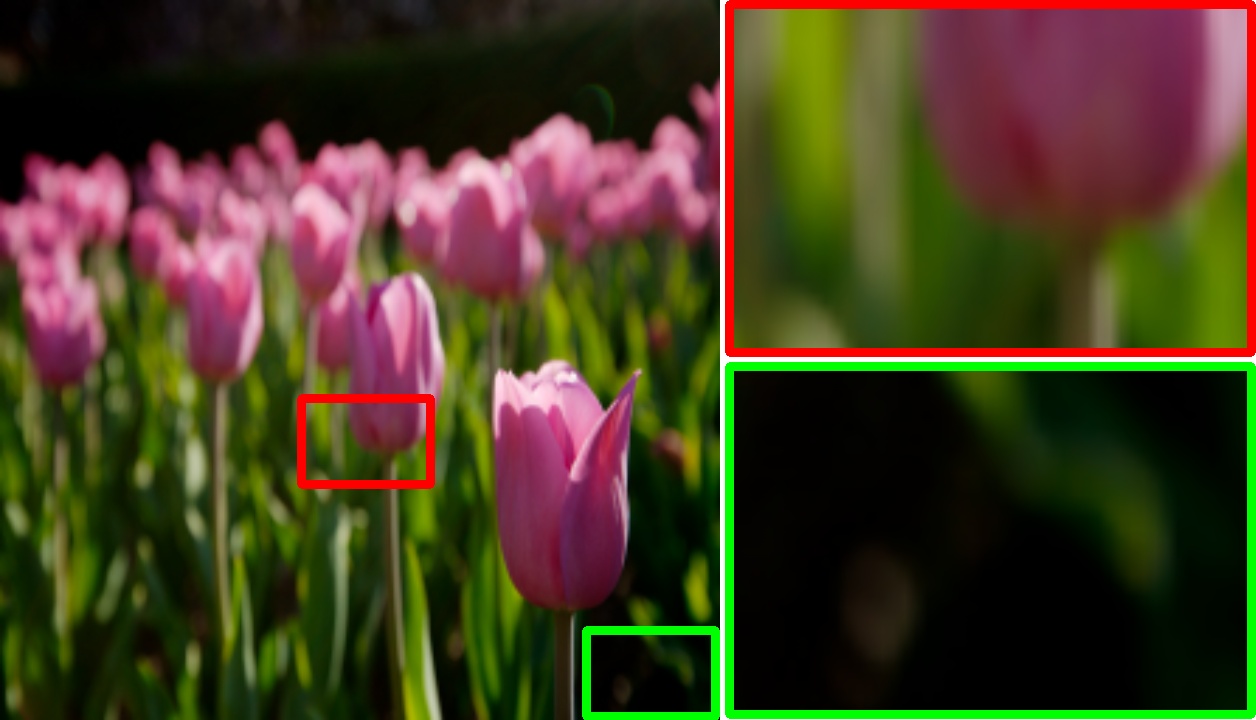}}
            \centerline{(h) Target}\medskip
        \end{minipage}
    \end{minipage}

    \caption{
    A visual comparison of different LLIE methods is conducted on the MIT-Adobe FiveK dataset. (b), (e), and (f) exhibit overexposure in the entire image. Both (d) and (e) exhibit either blurriness or distortion, respectively. 
    }
    \label{fig:ref}
\end{figure*}

\subsection{Experiment Settings}

In order to ensure the highest level of fidelity and reliability in our experiments, we have made the deliberate decision to exclusively use the VigSet dataset for vignetting removal. 
Moreover, we also conduct experiments in the field of LLIE to demonstrate the versatility of DeVigNet in global illuminance adjustment. 

In our experiments, we leverage the following datasets:

\textbf{VigSet}: This dataset contains 983 image pairs. We allocate 803 pairs for training purposes and reserve the remaining 180 pairs for testing.

\textbf{MIT-Adobe FiveK}~\cite{bychkovsky2011learning} \& \textbf{LOL-v1}~\cite{wei2018deep}: For these datasets, we adopt the experimental settings delineated in~\cite{wang2023ultra}, ensuring consistency with prior research.

\subsubsection{Evaluation Metrics}
We evaluate the quality of the enhanced images using four widely adopted metrics: Peak Signal-to-Noise Ratio (PSNR), Structural Similarity (SSIM), Mean Absolute Error (MAE). In addition, to better align with human perceptual conditions, we utilize the Learned Perceptual Image Patch Similarity (LPIPS). These metrics are commonly used in the assessment of image enhancement and low-level computer vision tasks. Higher PSNR and SSIM scores indicate improved visual quality, while lower MAE and LPIPS suggest better accuracy in representing the original image.

\subsubsection{Implementation Details}
We employ PyTorch to implement our model and conduct experiments using the NVIDIA A40 GPU. We utilize the Adam optimizer, adhering to its default parameter settings. During the training process, all models are trained at a resolution of $512\times512$ and subsequently tested at different resolutions. Additionally, we set the batch size to 1 and the learning rate to $1e-4$.

\subsection{Comparisons with State-of-the-Art}
We have performed a comprehensive evaluation of state-of-the-art vignetting removal methods on the VigSet dataset. The compared vignetting methods include SIVC~\cite{zheng2008single} and RIVC~\cite{lopez2015revisiting}. These methods serve as benchmarks for evaluating the performance of our proposed vignetting removal technique.

Furthermore, we also conduct additional experiments with state-of-the-art LLIE methods. We compare DeVigNet with traditional LLIE methods such as LIME~\cite{guo2016lime}, MSRCR~\cite{597272}, and NPE ~\cite{6512558}. Additionally, we evaluate the performance of learning-based LLIE methods, including WV-SRIE~\cite{fu2016weighted}, PM-SRIE~\cite{fu2015probabilistic}, Uformer~\cite{wang2021uformer}, KID~\cite{zhang2019kindling}, DSLR~\cite{lim2020dslr}, ELGAN~\cite{jiang2021enlightengan}, JieP~\cite{cai2017joint}, RUAS~\cite{liu2021retinex}, Zero-DCE~\cite{guo2020zero}, and Zero-DCE++~\cite{li2021learning}.

Qualitative results, as shown in Figure~\ref{fig:vigset} demonstrate that our method exhibits significant superiority over others in the vignetting dataset. Our results display enhanced image clarity and more prominent vignetting removal. Moreover, our method achieves competitiveness on the LLIE dataset as shown in Figure~\ref{fig:ref}.

In terms of quantitative results, by comparing our proposed method with these state-of-the-art vignetting and LLIE methods as displayed in Table~\ref{quan} and Table~\ref{tb:vigset}, we can assess its effectiveness and performance in relation to existing methods. This comprehensive evaluation provides insights into the strengths and weaknesses of different approaches and helps to advance the field of vignetting removal. 

Conventional methods heavily depend on assumptions based on physics, which are not consistently accurate. Additionally, current LLIE methods are not well-suited for addressing the issue of vignetting. Therefore, based on the data from Table~\ref{tb:vigset}, it is evident that traditional methods and LLIE methods exhibit substantial disparities compared to other methods in multiple metrics, particularly in terms of MAE. Additionally, we conducted supplementary quantitative experiments on a LLIE dataset, as presented in Table~\ref{quan}. The results illustrate the strong performance of our method in LLIE. Besides, with the advancement of current technology, capturing high-resolution images has become an essential routine. DeVigNet stands out as the most efficient approach in achieving superior results in vignetting removal across various resolutions when compared to other existing methods.

\subsection{Ablation Studies}
We maintain a similar parameter count across all combinations to ensure a fair comparison. DAFT and ACEM are removed respectively using the comment-out method in the experiments. As shown in Table~\ref{ablation}, we conduct ablation studies evaluating various components of our model at different image resolutions. These components include varying the depth of the Laplacian pyramid, as well as ablating the ACEM and DAFT modules.

As the depth of the Laplacian pyramid increases, the magnitude of the low-frequency component decreases. This reduction directly impacts the quality of vignetting removal, especially with respect to preserving low-frequency color information. 

DAFT plays a pivotal role in capturing global color information within the low-frequency component of an image. Therefore, models lacking DAFT suffer performance degradation due to the loss of global color context.

Meanwhile, ACEM is primarily utilized to retain edge details during image reconstruction, focusing on the high-frequency portion. Removing ACEM causes models to lose textural information in the reconstruction phase, thereby deteriorating performance due to the absence of high-frequency context.

Additionally, we replace DAFT with DHAN~\cite{2019Towards}, the results are shown in Table~\ref{ablation}. Accordingly, DAFT and ACEM are responsible for the low-frequency and high-frequency components, respectively. While each component contributes individually to improved vignetting removal performance, their combined effect yields optimal results by leveraging complementary global and local context.

\section{Conclusion}
In this paper, we introduce Vigset, the first large-scale high-resolution vignetting removal dataset with ground truth images. Vigset comprises 983 pairs of images captured under different lighting conditions and in various scenes. Additionally, we propose a novel method called DeVigNet, specifically designed for vignetting removal on this dataset. It includes three components: The Dual Aggregated Fusion Transformer, the Adaptive Channel Expansion Module and the Hierarchical Channel Attention Module. By utilizing the Laplacian pyramid, DeVigNet performs vignetting removal on the color information in the high-frequency and low-frequency domains of the image, thereby achieving optimal results. DeVigNet effectively eliminates vignetting effects in images, demonstrating superior performance compared to existing methods in terms of both quality and quantity for vignetting removal.

\section{Acknowledgement}
This work was supported in part by the Science and Technology Development Fund, Macau SAR, under Grant 0087/2020/A2 and Grant 0141/2023/RIA2, in part by the National Natural Science Foundations of China under Grant 62172403, in part by the Distinguished Young Scholars Fund of Guangdong under Grant 2021B1515020019, in part by the Excellent Young Scholars of Shenzhen under Grant RCYX20200714114641211.

\bibliography{aaai24}
\end{document}